\documentclass{sigchi-ext}
\usepackage[T1]{fontenc}
\usepackage{url}
\usepackage{amsmath}
\usepackage{textcomp}
\usepackage[scaled=.92]{helvet} 
\usepackage{graphicx} 
\usepackage{balance}  
\usepackage{booktabs} 
\usepackage{ragged2e} 
\usepackage{multirow}
\usepackage{subfigure}
\usepackage{url}
\usepackage{relsize} 

\copyrightinfo{Permission to make digital or hard copies of part or all of this work for personal or classroom use is granted without fee provided that copies are not made or distributed for profit or commercial advantage and that copies bear this notice and the full citation on the first page. Copyrights for third-party components of this work must be honored. For all other uses, contact the Owner/Author. \\
Copyright is held by the owner/author(s).\\
{\emph{CSCW'16 Companion}}, February 27 - March 02, 2016, San Francisco, CA, USA \\
ACM 978-1-4503-3950-6/16/02. \\
http://dx.doi.org/10.1145/2818052.2869110}
\title{WASSUP? LOL : Characterizing Out-of-Vocabulary Words in Twitter}

\numberofauthors{7}
\author{
\alignauthor{\textbf{Suman Kalyan Maity}\\
       \affaddr{Dept. of CSE}\\
       \affaddr{IIT Kharagpur, India}}
       \alignauthor{\textbf{Chaitanya Sarda}\\
       \affaddr{Dept. of CSE}\\
       \affaddr{IIT Kharagpur, India}}
       \vfil \alignauthor{\textbf{Anshit Chaudhary}\\
       \affaddr{Dept. of CSE}\\
       \affaddr{IIT Kharagpur, India}}
       \alignauthor{\textbf{Abhijeet Patil}\\
       \affaddr{Dept. of CSE}\\
       \affaddr{IIT Kharagpur, India}}
       \vfil \alignauthor{\textbf{Shraman Kumar}\\
       \affaddr{Dept. of CSE}\\
       \affaddr{IIT Kharagpur, India}}
       \alignauthor{\textbf{Akash Mondal}\\
       \affaddr{Dept. of CSE}\\
       \affaddr{IIT Kharagpur, India}}
       \vfil \alignauthor{\textbf{Animesh Mukherjee}\\
       \affaddr{Dept. of CSE}\\
       \affaddr{IIT Kharagpur, India}}}

\def\plaintitle{WASSUP? LOL : Characterizing Out-of-Vocabulary Words in Twitter} \def\plainauthor{First Author, Second Author, Third Author,
  Fourth Author, Fifth Author, Sixth Author}
\def\plainkeywords{Authors' choice; of terms; separated; by
  semicolons; include commas, within terms only; required.}

\hypersetup{%
  pdftitle={\plaintitle}, pdfauthor={\plainauthor},
  pdfkeywords={\plainkeywords}, }

\begin{document}

\maketitle

\RaggedRight{} 

\begin{abstract}
Language in social media is mostly driven by new words and spellings that are constantly entering the lexicon thereby polluting it and resulting in high deviation from the formal written version. The primary entities of such language are the out-of-vocabulary (OOV) words. In this paper, we study various sociolinguistic properties of the OOV words and propose a classification model to categorize them into at least six categories. We achieve 81.26\% accuracy with high precision and recall. We observe that the content features are the most discriminative ones followed by lexical and context features.
\end{abstract}

%
\category{J.4}{Computer Applications}[Social and Behavioral Sciences]

\section{Introduction}
From ``unfollow'' to ``selfie'', the rise of social media has a significant impact on our language. The language we see/use in social media is very different from standard written language. Frequent alterations, abbreviations, acronyms, neologisms have left our language in a continuous state of alteration and regeneration. These social media specific forms have made the language ``bad'' and does not meet our expectations about vocabulary, spelling and syntax~\cite{dan}. However, these forms have accelerated the real-time typed conversation; on mobile phones they have minimized the inconvenience of typing with tiny keys and on Twitter they have helped us packing our thoughts/feelings within 140 characters. Not only that, the social media language is very rich with a wide range of linguistic innovations including emoticons, abbreviations, shortenings, lengthenings etc~\cite{bad}. 
\begin{margintable}[1pc]
  \begin{minipage}{\marginparwidth}
    \centering
\small
\captionsetup{font=small}
\caption{Example words in each categories.}
\label{one}
\begin{tabular}{|p{2cm}|p{2cm}|}\hline
OOV Categories & Examples\\\hline
Emoticons  &  :), :(, :D, :P, :/\\ \hline
Word Lengthenings & noooo, pleaseeee, okk, damnnn\\ \hline
Expressions & haha, uhh, ughh, ahah, grr\\ \hline
Word Shortenings + Abbreviations& lol, omg, yolo, rofl, oomf\\ \hline
Proper Nouns & instagram, miley, bieber, mcdonalds, tumblr\\ \hline
Word Mergings & wassup, iknow, followback \\ \hline
\end{tabular}
\end{minipage}
\end{margintable}
\begin{marginfigure}[1pc]
  \begin{minipage}{\marginparwidth}
    \centering
\includegraphics*[scale = 0.35,angle=0]{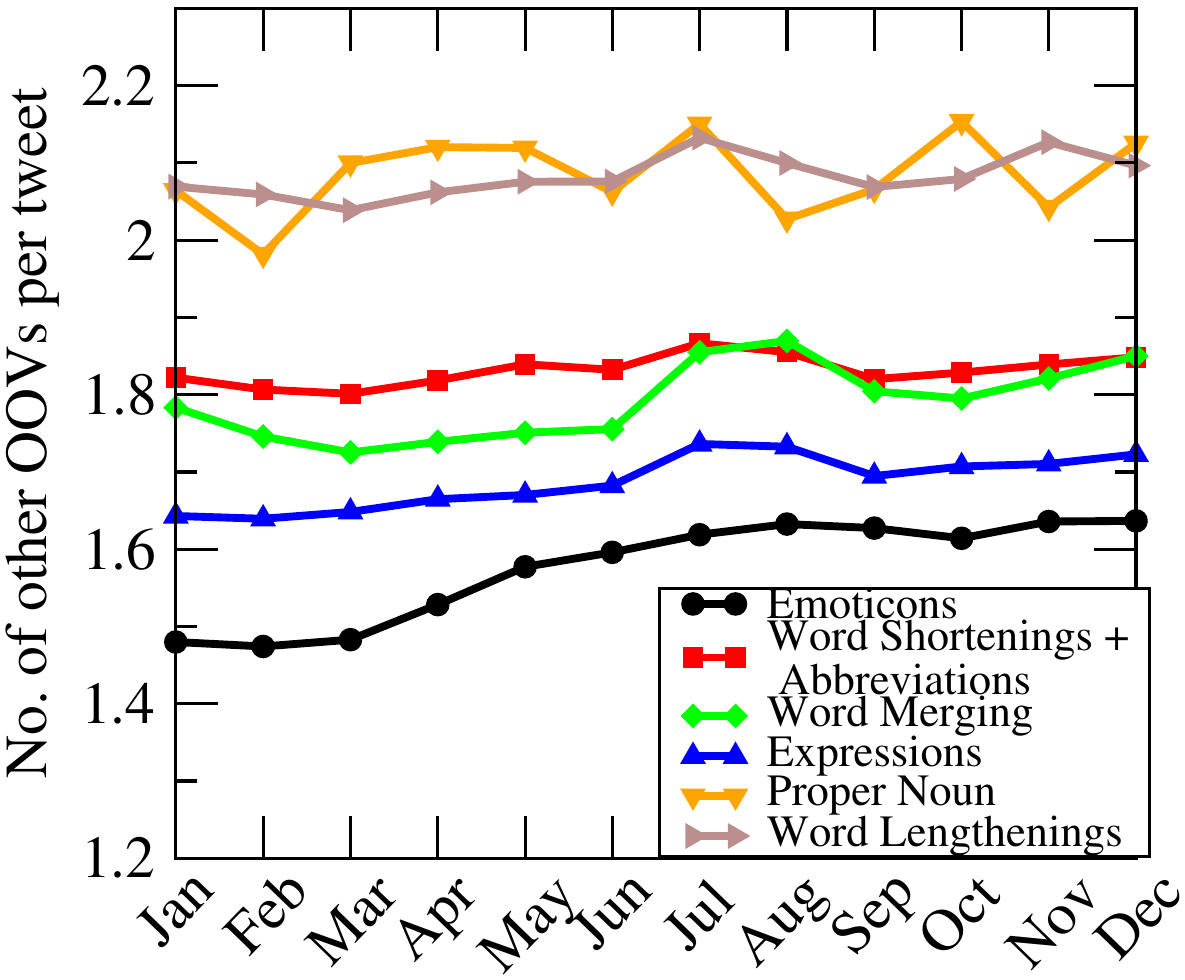}
\caption{\label{fig1} Category specific monthwise distribution of other co-occurring OOV words per tweet for 2013.}
\end{minipage}
\end{marginfigure}
In this work, we categorize these out-of-vocabulary (OOV) words into at least six different interesting categories and subsequently propose an automatic method to perform such a categorization. This categorization is useful in further interpretation of the semantics of each of the categories. So far the common practice has been to dispose the OOV words before any further processing or application development; an accurate semantic interpretation might save this rich dataset from disposal and increase the possibility of using it more meaningfully.
\section{Dataset description}
We collected Twitter $1\%$ random sample of tweets from $1^{st}$ July, 2011 to $31^{st}$ December, 2013. For analysis, we consider users who have mentioned English as their language in their profile. We also performed a second level filtering of the tweets by a language detection software\footnote{\url{https://github.com/saffsd/langid.py}} to remove any non-English tweets from the dataset. The data is then tokenized using the same tokenizer used by the CMU POS tagger~\cite{owu}. In total, the dataset consists of $\sim1$ billion tweets. 
\section{Categorization of OOV words} We use GNU Aspell dictionary\footnote{\url{http://aspell.net/}} to find out the OOV words from the set of tweets. We then select top 3500 stable OOV words that occur in all the months over the span of 2.5 years and manually identify 6 major categories for these OOV words: emoticons, expressions, word shortenings + abbreviations, proper nouns, word lengthenings, word mergings. Five of the authors separately labeled the words. The inter-annotator agreement is found to be very high (Fleiss' $\kappa$ = $0.96$). Out of 3500 words, we attained consensus on 3259 words. The remaining ambiguous words are not taken into consideration for further study. Table~\ref{one} shows example OOV words belonging to each of these categories. In Fig~\ref{fig1}, we show the monthwise distribution of other OOV words per tweet appearing with a particular OOV word for individual categories. We observe a clear distinction among most of the categories - the tweets containing ``emoticons'' have the 
lowest content of other OOV words per tweet whereas for the ``proper noun'' class, the other OOV content per tweet is the highest.

Among the above six OOV categories, some of them are relatively easy to distinguish. Words in emoticons and word lengthenings category are made up of very regular building blocks. For example, emoticons are made up of punctuation symbols. Using simple regular expressions for presence of punctuation symbols, we could achieve a very high accuracy of 98.1\% (Precision: 0.877, Recall: 0.976). Word lengthenings are essentially created by repeating one or more letters in the word. In such case, we first find the repetitive letters and then remove them one by one and check with the dictionary to obtain the corresponding dictionary word. This method yields an accuracy of 93.1\%, however the precision and recall rates achieved are 0.432 and 0.677 respectively. It is not easy to come up with proper regular expressions that can distinguish the other categories. It is really difficult to disambiguate the words like LOL, ROFL, WASSUP etc. by just writing a regular expression. This is one of the primary motivations behind our feature-based automatic categorization framework.

\section{Automatic classification of OOV words}
In this section, we propose a classification model for automatic categorization of OOV words into one of the last four categories in Table~\ref{one}. For the task of classification, we consider a random sample of 5000 tweets for each of the OOV words and then learn three major types of features : \\
\begin{margintable}[1pc]
  \begin{minipage}{\marginparwidth}
    \centering
\small
\captionsetup{font=small}
\caption{Performance of various classifier for different topic selection for LDA feature with number of topics ($K$ = 10, 20, 30, 40, 50). First five lines refer to results for SVM classifier and last five for Logistic Regression.}
\label{two}
 \resizebox{4.9cm}{!}{\begin{tabular}{|p{0.2cm}|p{0.6cm}|p{0.6cm}|p{0.6cm}|p{0.6cm}|p{0.6cm}| }
\hline
 $K$ & Accur-acy &Preci-sion & Recall & F-Score & ROC Area \\ \hline
10 & 80.09 & 0.797 & 0.801 & 0.796 & 0.845 \\ \hline
20 & 79.85 & 0.795 & 0.798 & 0.794 & 0.844 \\ \hline
30 & 80.53 & 0.801 & 0.805 & 0.8 & 0.848 \\ \hline
40 & 79.73 & 0.793 & 0.797 & 0.793 & 0.843 \\ \hline
50 & 81.26 & 0.81 & 0.813 & 0.809 & 0.855 \\ \hline \hline
10 & 79.81 & 0.794 & 0.798 & 0.793 & 0.922 \\ \hline
20 & 78.92 & 0.784 & 0.789 & 0.784 & 0.923 \\ \hline
30 & 80.09 & 0.796 & 0.801 & 0.796 & 0.922 \\ \hline
40 & 79.52 & 0.791 & 0.795 & 0.79 & 0.92\\ \hline
50 & 80.21 & 0.797 & 0.802 & 0.796 & 0.925 \\ \hline
  \end{tabular}}
\end{minipage}
\end{margintable}
\begin{margintable}[1pc]
  \begin{minipage}{\marginparwidth}
    \centering
    \small
\captionsetup{font=small}
\caption{Performance of various combinations of feature categories for $K=50$.}
  \label{feature_combo}
 \begin{tabular}{|p{2.5cm}|p{1.2cm}|}
\hline
\textbf{All} & \textbf{81.26\%}\\ \hline
 Lexical + Content & 80.38\%  \\ \hline
 Content + Context & 79.12\% \\ \hline
 Lexical + Context & 74.48\%  \\ \hline
 \textbf{Content} & \textbf{77.83\%} \\ \hline
 Lexical & 72.37\% \\ \hline
 Context & 58.71\% \\ \hline
  \end{tabular}
\end{minipage}
\end{margintable}
\textbf{Lexical features} - features related to the lexical properties of the words around OOV words.\\
\textbf{Content features} - features related to the content of the tweets in which the OOV words appear. \\
\textbf{Context features} - these include positioning and placement information of various entities in the tweets.\\
\textbf{Lexical features:}\\
\textbf{Distribution of POS tags of the words appearing with the OOV word:} We use standard CMU POS tagger~\cite{owu} for identifying the POS tags of the words in the tweets. We then consider fraction of words belonging to each of the 21 POS tag categories as features.\\
\textbf{Part-of-speech diversity of the words surrounding an OOV word:} We define the POS diversity (POSDiv) for an OOV word as the entropy of the probability $p_j$ of one of its surrounding word to have the $j^{th}$ POS in the set of POS tags. We use this diversity metric as a feature for our classifier.\\
\textbf{Distribution of named entities (NE) appearing with OOV word:} We also perform named entity recognition~\cite{ritter} of the words appearing in the tweets to understand which types of entities co-occur with which type of OOV word. We consider fraction of words belonging to each of the NE categories as a feature for the classification model.\\
\textbf{Content features:}\\
\textbf{Length of the OOV word:} The length of an OOV word is significant in determining the category it belongs. The abbreviations, shortenings are generally of smaller length whereas expressions, word lengthenings and word merging are usually of longer length.\\
\textbf{Word diversity:} This feature tells us how much diverse are the words related to an OOV. If $D_i$ is the document containing all the tweets in which OOV $O_i$ appears and $p(w|D_i)$ is the probability of a word belonging to the document $D_i$ then word diversity of hashtag $O_i$ is the entropy of $p(w|D_i)$.\\
\textbf{Avg. hashtag clarity:} Hashtag clarity, a metric defined in~\cite{sun} quantifies topical cohesiveness of all the tweets in which the hashtag appears. The clarity of the hashtags co-occurring with an OOV word constitutes a feature for the classification model.\\
\textbf{Presence of Twitter entities appearing with the OOV word:} We consider appearances of various Twitter entities (no. of hashtags, no. of mentions, no. of retweets) in the tweets as features discriminating the various categories of OOV.\\
\textbf{Distribution of topics:} For topic discovery from the tweet corpus, we adopt the Latent Dirichlet Allocation (LDA)~\cite{lda} model. For an OOV $O_i$, we consider all the 5000 tweets in which the OOV appears as a document (no. of topics as $K$ = 10, 20, 30, 40, 50) for the LDA model and find out $p(topic_k|D_i)$ for a document $D_i$ containing all the tweets in which the $i^{th}$ OOV appears. Each of these $p(topic_k|D_i)$ act as a feature to the model.\\
\textbf{Topical diversity:} We also compute topical diversity ($TopicDiv$) of an OOV ($O_i$) from the document-topic distributions obtained above as the entropy of the probability $p(topic_k|D_i)$.\\
\textbf{Cognitive dimension:} Words from a document containing all the OOVs are classified into various linguistic and psychological categories by LIWC software~\cite{liwc}. We consider 42 such classificatory features.\\
\textbf{Context features:} \\
\marginpar{%
  \vspace{-15pt} \fbox{%
    \begin{minipage}{0.925\marginparwidth}
\textbf{Performance evaluation of the classifier:} We have used SVM and logistic regression classifiers for classification. We perform a 10-fold cross-validation and achieve 81.26\%\footnote{Even if we take a random sample of OOV words considering as less as 1000 tweets each, we achieve $\sim$72\% accuracy. This result is relevant for the rarer OOV words} accuracy with high precision and recall rates (see table~\ref{two} for details). Both the classifiers yield very similar classification performance; however the logistic regression model gives better area under the ROC curve compared to SVM. Number of topics ($K$) of LDA do not have a significant effect on the classification results. We observe that content features are the strongest feature type whereas context features are relatively weak (see table~\ref{feature_combo}). For individual features, we rank them by descending order of chi-square ($\chi^2$) value and observe that length of the OOV, LIWC features like ingestion (biological process), assent (spoken category),
 LDA topical features and topical diversity feature are the most discriminative ones.\end{minipage}}\label{sec:performance} }
\textbf{Distribution of other OOV words around the OOV:} Presence of other OOV words around an OOV is an important aspect. We consider the fraction of tweets containing 0, 1, 2 or more other OOVs as four classificatory features. We also consider average number of other OOVs in tweets as feature.\\
\textbf{Placement of other OOV words and IV words in close proximity around the OOV:} This feature captures the context of an OOV ($O$) in the tweets. We consider the fraction of tweets having other OOVs at a distance 1, 2, greater than 3 from the position of $O$ as three classificatory features. We also consider the fraction of tweets having IV words at a distance 1, 2, greater than 3 from the position of $O$ as features in the model.\\
\textbf{Position of the OOV word in the tweets:} Expressions generally appear at the end of the tweet while the abbreviations are usually positioned in the middle. For identifying position of a word in a tweet, we consider normalized position i.e, no. of words positioned before it, divided by total no of words. We have 3 positions for a word based on average normalized position value : $<0.3$, $0.3-0.7$, $>0.7$ for left, middle and right respectively.\\
\textbf{Position of the Twitter entities in the tweets:} As stated above, we consider average normalized position of hashtags and mentions in the tweets as features of the model.
\section{Conclusions}
In this paper, we have investigated various categories of OOV words and proposed an automatic method to classify them into six disjoint classes. Our proposed classification framework achieves a high accuracy of 81.26\% with high precision and recall. We observe that content features are the most discriminative which alone contribute to accuracy of 77.83\%. This leads us to believe that there are strong semantic differences in various categories of OOV words which could be used for identification of them and would also be very effective in classifying so far unseen OOV words that get introduced in the social media platform very frequently. This classification technique could also be useful as stepping stone toward improving accuracy of the text processing algorithms on social media texts and other linguistic analysis - e.g., if we can use this algorithm to identify the combined words well, we can use techniques that split those words into their components for other analysis. Linguists may also be interested in tracking how these 6 categories change over time, or how people in different regions or different age groups use them in differing proportions.  


\bibliographystyle{SIGCHI-Reference-Format}
\bibliography{ref}

\end{document}